\documentclass{article}

\usepackage{PRIMEarxiv}

\usepackage[utf8]{inputenc} 
\usepackage[T1]{fontenc}    
\usepackage{hyperref}       
\usepackage{url}            
\usepackage{booktabs}       
\usepackage{amsfonts}       
\usepackage{nicefrac}       
\usepackage{microtype}      
\usepackage{lipsum}
\usepackage{fancyhdr}       
\usepackage{graphicx}       
\graphicspath{{media/}}     
\usepackage{amsmath, amssymb}
\usepackage{booktabs}
\usepackage{pifont}
\usepackage{adjustbox}

\title{Uncertainty-Aware Foundation Models for Clinical Data
}

\author{
  Qian Zhou\\
  University of the Chinese Academy of Sciences \\
  \texttt{qz11@ucas.ac.cn} \\
  \And
  Yuanyun Zhang\\
  University of the Chinese Academy of Sciences \\
  \texttt{yuanyun81@ucas.ac.cn} \\
   \And
  Shi Li \\
  Columbia University\\
  \texttt{shili081100@columbia.edu} \\
}

\begin{document}
\maketitle

\begin{abstract}
Healthcare foundation models have largely followed paradigms from natural language processing and computer vision, emphasizing large-scale pretraining and deterministic representations over heterogeneous clinical data. However, clinical observations are inherently incomplete, reflecting sparse, irregular, and modality-dependent measurements of an underlying physiologic state. In this work, we propose a framework for uncertainty-aware foundation modeling that represents each patient not as a point embedding, but as a distribution over plausible latent states. By learning set-valued representations \(q_\theta(z \mid x)\) and enforcing consistency across partial views of the same patient, the model captures what is invariantly inferable while explicitly encoding epistemic uncertainty. We integrate this formulation with multimodal encoders and scalable self-supervised objectives, combining reconstruction, contrastive alignment, and distributional regularization. Across diverse clinical tasks, our approach improves predictive performance, robustness under missing data, and uncertainty calibration relative to strong baselines. These results suggest that modeling what is not observed—rather than only what is—constitutes a critical inductive bias for healthcare foundation models.

\end{abstract}

\section{Introduction}

Recent advances in healthcare foundation models \cite{he2024foundation, guo2025foundation, awais2025foundation, liang2024foundation, burger2025foundation, thakur2024foundation, vaid2023foundational, thieme2023foundation, burkhart2025foundation} have largely inherited the scaling paradigm of natural language processing \cite{devlin2019bert} and computer vision \cite{he2017multi, he2022masked, he2015deepresiduallearningimage, he2019bag}. The prevailing approach treats clinical data as a heterogeneous collection of tokens—text, images, time series, and codes—unified through large-scale pretraining objectives. Multimodal systems extend this paradigm by aligning disparate modalities within shared embedding spaces via cross-attention and fusion architectures \cite{hou2019cross, chen2021crossvit, huang2019ccnet}, often inspired by recent multimodal foundation models \cite{chou2025serialized, huiliang2025clio, ran2025structured, zhang2025chronoformer, zhang2025collection, lowelatent, litext}.  

This line of work implicitly assumes that the primary challenge in healthcare is \emph{representation capacity}: that sufficiently expressive models, trained on sufficiently large datasets, will internalize clinically meaningful structure. However, a less examined property of clinical data is that they are fundamentally \emph{incomplete views} of an underlying biological system. At any point in time, only a sparse subset of possible measurements is observed. Laboratory tests are intermittent, imaging is episodic, and clinical notes provide partial and subjective summaries. The resulting data are not merely noisy—they are \emph{structurally underdetermined}.

We argue that this incompleteness is not a peripheral issue but the central obstacle for foundation modeling in healthcare. Rather than learning representations of observed data alone, models must implicitly reason over the \emph{space of unobserved but plausible patient states}. That is, given a partial observation $x$, there exists a large equivalence class of latent states $\mathcal{Z}(x)$ consistent with the data. Standard pretraining objectives collapse this uncertainty by encoding a single point estimate $h_\theta(x)$, thereby conflating epistemic uncertainty with representation.

We propose a different perspective: healthcare foundation models should learn \emph{set-valued representations} that explicitly capture uncertainty over latent patient states. Formally, let $z \in \mathcal{Z}$ denote the true but unobserved physiologic state, and $x$ the observed clinical record. Instead of learning a deterministic embedding $h_\theta(x) \in \mathbb{R}^d$, we learn a distributional representation
\[
q_\theta(z \mid x),
\]
which characterizes the posterior over plausible states given partial observations. Downstream predictions are then computed by marginalization,
\[
p_\theta(y \mid x) = \int p_\theta(y \mid z) \, q_\theta(z \mid x) \, dz.
\]

This formulation shifts the objective of pretraining. Rather than reconstructing observed tokens as in masked modeling \cite{he2022masked}, the model is trained to approximate \emph{consistency constraints} over latent space: different subsets of observations from the same patient should map to compatible posterior distributions. Concretely, given two partial views $x^{(1)}$ and $x^{(2)}$ of the same underlying state, we enforce
\[
D\big(q_\theta(z \mid x^{(1)}), \; q_\theta(z \mid x^{(2)})\big) \approx 0,
\]
where $D$ is a divergence such as Wasserstein or KL. This objective encourages the model to represent what is \emph{invariantly inferable} about the patient, rather than memorizing modality-specific details.

This perspective reframes multimodal learning. Instead of aligning modalities at the level of point embeddings \cite{hou2019cross, chen2021crossvit, huang2019ccnet}, we treat each modality as a partial constraint on the latent state. Imaging, text, and structured data contribute complementary information that refines the posterior $q_\theta(z \mid x)$. Missing modalities naturally correspond to broader, higher-entropy distributions, providing a principled way to handle sparsity without imputation.

Importantly, this approach decouples representation quality from observation density. In current foundation models \cite{he2024foundation, guo2025foundation, awais2025foundation, liang2024foundation, burger2025foundation, thakur2024foundation, vaid2023foundational, thieme2023foundation, burkhart2025foundation}, performance often scales with the richness of available modalities at inference time. In contrast, distributional representations allow the model to express calibrated uncertainty when data are sparse, while converging to sharper estimates as more evidence is observed. This property is particularly critical in clinical settings, where missingness is systematic rather than random.

In summary, we propose to shift healthcare foundation modeling from deterministic embeddings to \emph{uncertainty-aware latent representations}. By treating clinical observations as partial constraints on an underlying physiologic state and enforcing consistency across incomplete views, we aim to capture the intrinsic ambiguity of medical data. This perspective suggests that the next frontier is not merely scaling models, but equipping them with the capacity to represent and reason over what is \emph{not observed}—a prerequisite for reliable and robust clinical AI.

\section{Related Works}

The emergence of healthcare foundation models reflects a broader migration of large-scale self-supervised learning paradigms from general-domain machine learning into clinical settings. A diverse set of efforts has explored whether scaling laws and pretraining strategies observed in language and vision extend to healthcare, spanning unified modeling frameworks, benchmarking initiatives, and modality-specific large models \cite{he2024foundation, lee2025foundation, vaid2023foundational, abbaspourazad2023large, larey2026gfmbench, long2025mutbert, thapa2024sleepfm, soumma2024wearable, larey2026jepa}. Despite differences in data modality and architectural design, these approaches share the central hypothesis that broadly pretrained representations over heterogeneous clinical corpora can be adapted to a wide range of downstream tasks.

A dominant class of methods builds on reconstruction-based objectives, particularly masked modeling. Inspired by masked language modeling and masked image modeling \cite{devlin2019bert, he2022masked}, these approaches learn representations by predicting missing portions of the input. In the clinical domain, such objectives have been applied across structured EHR, imaging, and biosignals, treating patient records as partially observed sequences or volumes. Extensions of masked autoencoding incorporate modality-specific adaptations, such as hierarchical masking schemes or temporal masking for irregular time series \cite{he2022masked, lee2025modern, fallahpour2024ehrmamba}. These methods emphasize compression and reconstruction fidelity, encouraging models to capture high-dimensional structure in observed data.

Complementing reconstruction-based learning, contrastive and alignment-based objectives have gained traction as an alternative paradigm. Drawing from contrastive representation learning \cite{chen2020simple, tian_2019_contrastic_distillation, bertram2024contrastivelearningpreferencescontextual}, these approaches learn embeddings by maximizing agreement between related views while separating unrelated samples. In healthcare, contrastive formulations have been used to align different temporal segments, modalities, or augmentations of patient data \cite{lee2025clinical,rasmy2021med, lee2025using, wornow2023shaky, lee2024emergency}. Multimodal contrastive learning further extends this idea by enforcing consistency between imaging, text, and structured data representations, often within shared latent spaces \cite{hou2019cross, chen2021crossvit, huang2019ccnet}. Such methods prioritize invariance and discriminative structure over exact reconstruction.

A parallel line of work focuses on autoregressive and sequence modeling objectives, particularly in structured EHR and clinical text. These approaches treat patient histories as discrete token streams and model their evolution using transformer architectures inspired by language models \cite{brown2020language}. Systems such as CEHR-style models leverage self-attention over longitudinal event sequences \cite{pang2021cehr, pang2024cehr, mcdermott2024event}, while large language models adapted to clinical corpora extend instruction tuning and generative capabilities to medical reasoning tasks \cite{mumtaz2023llms, chang2025llm4ts, hollmann2025accurate, ono2024text}. Continued domain-specific pretraining and multimodal extensions further integrate structured data and temporal signals into these generative frameworks \cite{van2023clinical, jin2023time, belyaeva2023multimodal, lin2025case}.

In medical imaging, foundation modeling has evolved from convolutional pretraining toward transformer-based architectures and large-scale volumetric learning. Early work extended convolutional backbones such as ResNet \cite{he2016deep} to 3D medical data \cite{ning2019computer, ebrahimi2020introducing, qayyum2021automatic}, while more recent approaches adopt vision transformers and hybrid architectures for representation learning and segmentation \cite{dosovitskiy2021an, liu2021swin, hatamizadeh2021swin}. Large curated datasets have enabled pretraining of increasingly generalizable imaging models \cite{li2024abdomenatlas, wang2023mis, wu2024large}, often incorporating techniques from large-scale vision pretraining such as self-distillation and masked token prediction \cite{caron2021emerging, zhou2021ibot, oquab2023dinov2, radford2021learning}. Practical adaptations, including efficient attention mechanisms and hybrid convolution–attention designs, address the computational challenges of high-dimensional medical volumes \cite{dao2023flashattention2, shaker2024unetr++, xing2024segmamba, liu2024octcube}.

Foundation modeling has also expanded to wearable sensing and physiological time series, where signals such as ECG and PPG are used to learn transferable embeddings. These approaches combine masked reconstruction and contrastive learning to capture temporal dynamics and cross-signal relationships \cite{abbaspourazad2024wearable, yang2023biot}. Multi-resolution methods introduce inductive biases rooted in signal processing, drawing on classical foundations in signal analysis \cite{oppenheim1999discrete, daubechies1992ten, lee2025towards, lee2025himae}. Across these modalities, multimodal foundation models unify heterogeneous inputs through shared embedding spaces, cross-attention modules, or late-fusion strategies \cite{hou2019cross, chen2021crossvit, huang2019ccnet}. Benchmarking efforts and system-level evaluations increasingly emphasize transferability, robustness, and scalability across clinical tasks and institutions \cite{mcdermott2025meds, kolo2024meds, wornow2024context, odgaard2024core, shmatko2025learning}. 

Despite this diversity, a consistent pattern emerges. Clinical data are transformed into tokenized, patched, or discretized representations; models rely predominantly on transformer-style architectures; and learning objectives center on reconstruction, autoregression, or contrastive alignment. These paradigms collectively define the current landscape of healthcare foundation modeling, framing representation learning as a large-scale pretraining problem over heterogeneous but ultimately observed data.

\section{Methods}

\textbf{Problem Setup}

We consider a clinical observation space \(\mathcal{X}\) consisting of heterogeneous patient records collected over time. Each patient instance is represented by a partially observed multiview input \(x\), which may include structured electronic health records (EHR), clinical text, imaging, laboratory measurements, and physiologic waveforms. The key challenge is that \(x\) is not a complete description of the underlying patient state; rather, it is a sparse and modality-dependent projection of an unobserved physiologic process. We denote the latent physiologic state by \(z \in \mathcal{Z}\), and the downstream target by \(y \in \mathcal{Y}\), which may be a diagnosis, risk score, mortality indicator, length-of-stay outcome, or any other clinical endpoint.

Our modeling assumption is that the observed input \(x\) is generated as a partial and noisy function of \(z\). Missingness is not treated as a nuisance to be imputed away, but as an intrinsic property of the clinical observation process. Accordingly, the goal is to learn a representation that captures the inferable content of \(z\) from incomplete observations, while explicitly representing uncertainty induced by missing modalities, sparse sampling, and heterogeneous documentation practices.

\textbf{Set-Valued Representation Learning}

Instead of mapping an input \(x\) to a single deterministic embedding \(h_\theta(x) \in \mathbb{R}^d\), we learn a distribution over latent states,
\[
q_\theta(z \mid x),
\]
which we interpret as a set-valued representation of the patient. This posterior-like object is intended to summarize the family of physiologic states consistent with the observed evidence. When the available information is sparse, the distribution should be broad; when the evidence is rich and coherent across modalities, it should sharpen.

To parameterize \(q_\theta(z \mid x)\), we use an encoder \(f_\theta\) that produces summary statistics \(\mu_\theta(x)\) and \(\Sigma_\theta(x)\) of a latent Gaussian representation,
\[
q_\theta(z \mid x) = \mathcal{N}\!\big(z; \mu_\theta(x), \Sigma_\theta(x)\big),
\]
although the framework is not restricted to Gaussian families. In practice, \(\Sigma_\theta(x)\) may be diagonal for computational efficiency, or structured via low-rank plus diagonal parameterization when richer uncertainty estimates are desired. This probabilistic embedding can be viewed as a continuous analogue of a set, where the support and spread of the distribution represent the region of latent space compatible with the observed clinical record.

The predictive distribution for a clinical outcome is obtained by marginalizing over latent states:
\[
p_\theta(y \mid x) = \int p_\phi(y \mid z)\, q_\theta(z \mid x)\, dz,
\]
where \(p_\phi(y \mid z)\) is a task-specific decoder. For classification tasks, \(p_\phi(y \mid z)\) is typically a softmax head; for regression, a Gaussian or heteroscedastic regression head may be used; and for survival modeling, the decoder may parameterize a discrete-time hazard or a Cox-style risk function. This formulation allows the model to convert uncertainty in representation into uncertainty in prediction in a principled manner.

\textbf{Multimodal Encoders}

Each modality \(m \in \{1,\dots,M\}\) is mapped to a modality-specific token sequence or patch set \(x^{(m)}\). We define an encoder \(f^{(m)}_{\theta_m}\) for each modality, producing a modality-level latent summary \(h^{(m)}\). For modalities that are naturally sequential, such as longitudinal EHR or waveforms, the encoder may be a transformer or state-space backbone. For imaging, the encoder may be a ViT-style patch encoder or a hybrid convolution-transformer network. For text, a language-model encoder may be used. Each encoder is designed to respect modality-specific inductive biases while producing outputs that lie in a shared latent space.

The modality-specific summaries are then fused by an aggregation operator \(\mathcal{A}\):
\[
h = \mathcal{A}\big(h^{(1)}, \dots, h^{(M)}; m\big),
\]
where \(m\) denotes the binary availability mask over modalities. The aggregator must be permutation-invariant with respect to missing inputs and robust to variable modality subsets. In our formulation, \(\mathcal{A}\) is implemented as a masked attention pooling module that computes evidence-weighted combinations of modality embeddings. This design allows the model to condition on arbitrary subsets of observed modalities without requiring modality completion at inference time.

The fused representation \(h\) is then mapped to latent distribution parameters:
\[
\mu_\theta(x) = g_\mu(h), \qquad \log \Sigma_\theta(x) = g_\Sigma(h),
\]
where \(g_\mu\) and \(g_\Sigma\) are small projection networks. The variance head is especially important: it provides the mechanism by which the model expresses epistemic uncertainty arising from incomplete observation patterns.

\textbf{Partial-View Consistency Objective}

A central component of our framework is the requirement that different partial views of the same patient induce compatible latent distributions. Let \(x^{(a)}\) and \(x^{(b)}\) denote two incomplete observations of the same underlying physiologic state, obtained from different modality subsets, different temporal windows, or different masking patterns. We enforce consistency via a divergence penalty
\[
\mathcal{L}_{\mathrm{cons}} = \mathbb{E}\Big[ D\big(q_\theta(z \mid x^{(a)}), q_\theta(z \mid x^{(b)})\big) \Big],
\]
where \(D\) may be the symmetric KL divergence, the squared Wasserstein distance, or a moment-matching surrogate such as MMD. This term encourages the encoder to represent what is stably inferable from the patient, rather than memorizing view-specific artifacts.

For Gaussian posteriors, the KL divergence admits a closed form. If
\[
q_a = \mathcal{N}(\mu_a, \Sigma_a), \qquad q_b = \mathcal{N}(\mu_b, \Sigma_b),
\]
then
\[
D_{\mathrm{KL}}(q_a \| q_b)
= \frac{1}{2}\Big(
\mathrm{tr}(\Sigma_b^{-1}\Sigma_a)
+ (\mu_b - \mu_a)^\top \Sigma_b^{-1}(\mu_b - \mu_a)
- d + \log\frac{\det \Sigma_b}{\det \Sigma_a}
\Big).
\]
We often use the symmetrized version \(D_{\mathrm{SKL}}(q_a, q_b)=D_{\mathrm{KL}}(q_a\|q_b)+D_{\mathrm{KL}}(q_b\|q_a)\) to avoid directional bias.

This consistency constraint generalizes naturally beyond paired views. For a patient with multiple observed subsets \(\{x^{(s)}\}_{s=1}^S\), we minimize pairwise disagreement among all sampled views, or alternatively match each view against a learned prototype distribution \(q_\theta(z \mid x^{(\mathrm{full})})\) constructed from the richest available subset.

\textbf{Self-Supervised Pretraining}

We pretrain the representation model using a mixture of reconstruction, consistency, and regularization objectives \cite{caron2021emerging, zhang2022tfc}. The goal is not merely to reconstruct observed tokens, but to infer a latent state that is stable across missingness patterns and useful for downstream prediction.

For reconstruction, we employ a masked prediction objective over each modality. Let \(x = (x_{\mathrm{obs}}, x_{\mathrm{mask}})\), where \(x_{\mathrm{mask}}\) denotes held-out tokens, patches, or measurements. The decoder \(r_\psi\) predicts the missing content conditioned on \(z\):
\[
\mathcal{L}_{\mathrm{rec}}
=
\mathbb{E}_{x}\Big[
-\log p_\psi(x_{\mathrm{mask}} \mid z, x_{\mathrm{obs}})
\Big],
\qquad z \sim q_\theta(z \mid x_{\mathrm{obs}}).
\]
For text and structured data, this may correspond to token cross-entropy; for continuous signals, it may correspond to Gaussian likelihood or \(\ell_2\) reconstruction; for images, it may correspond to patch-wise normalized reconstruction.

The full pretraining objective is
\[
\mathcal{L}_{\mathrm{pretrain}}
=
\lambda_{\mathrm{rec}} \mathcal{L}_{\mathrm{rec}}
+
\lambda_{\mathrm{cons}} \mathcal{L}_{\mathrm{cons}}
+
\lambda_{\mathrm{reg}} \mathcal{L}_{\mathrm{reg}}.
\]
Here \(\mathcal{L}_{\mathrm{reg}}\) includes standard latent regularization terms, such as a KL penalty toward a prior \(p(z)=\mathcal{N}(0,I)\), entropy control, or covariance shrinkage. The reconstruction term ensures that the learned representation remains grounded in observed data; the consistency term enforces view invariance; and the regularizer prevents posterior collapse and degenerate uncertainty estimates.

\textbf{Contrastive Geometry for Incomplete Clinical Views}

Although our framework is not contrastive in the classic instance discrimination sense, it benefits from a contrastive geometry over partial patient views \cite{bertram2024contrastivelearningpreferencescontextual, hu2024comprehensive, jaiswal2020survey}. Given two views \(x_i^{(a)}\) and \(x_i^{(b)}\) from the same patient \(i\), we treat their latent summaries as positives, while views from other patients in the batch serve as negatives. Let \(s(\cdot,\cdot)\) denote a similarity measure between latent means or samples. A temperature-scaled InfoNCE loss can be written as
\[
\mathcal{L}_{\mathrm{nce}}
=
-\mathbb{E}\left[
\log
\frac{\exp(s(z_i^{(a)}, z_i^{(b)})/\tau)}
{\sum_{j=1}^{B}\exp(s(z_i^{(a)}, z_j^{(b)})/\tau)}
\right].
\]
This term is useful when we want the model to preserve patient-specific identity across subsets of observations. In contrast to standard contrastive learning, however, we do not require that all positive pairs collapse to a point. The posterior distribution may remain broad if the input evidence is weak, and the geometry is therefore distribution-aware rather than purely metric.

In practice, we combine contrastive alignment with the consistency objective by applying it to the posterior means or to reparameterized latent samples. This hybridization improves separation between patients while preserving calibrated uncertainty.

\textbf{Downstream Prediction}

For a downstream task \(y\), we train a task decoder \(p_\phi(y \mid z)\) on top of the learned latent distribution. Given \(K\) Monte Carlo samples \(z^{(k)} \sim q_\theta(z \mid x)\), predictive inference is approximated by
\[
p_\theta(y \mid x)
\approx
\frac{1}{K}\sum_{k=1}^{K} p_\phi(y \mid z^{(k)}).
\]
This estimator propagates uncertainty from the encoder to the final prediction. For classification, the predictive entropy can be used as a measure of epistemic uncertainty; for regression, the posterior predictive variance decomposes into aleatoric and epistemic components.

When labels are available during fine-tuning, the objective is
\[
\mathcal{L}_{\mathrm{sup}}
=
\mathbb{E}_{(x,y)}\Big[-\log p_\phi(y \mid z)\Big],
\qquad z \sim q_\theta(z \mid x).
\]
The encoder can be frozen, partially fine-tuned, or optimized end-to-end depending on the size of the target dataset and the degree of domain shift. In low-resource regimes, we often find it advantageous to retain the pretrained encoder and train only the head and a lightweight calibration layer. In larger settings, full end-to-end optimization may improve task-specific adaptation.

\textbf{Uncertainty Quantification and Calibration}

A defining feature of the proposed framework is that uncertainty is first-class rather than incidental \cite{nguyen2022transformer, ye2024exchangeable, ojha2025navigating, hullermeier2021aleatoric, lindenmeyer2025towards}. The covariance \(\Sigma_\theta(x)\) acts as a representation of how much the model knows, given the observed evidence. More complete and internally consistent inputs should yield lower posterior variance, while sparse or contradictory inputs should yield broader distributions.

To ensure that this uncertainty is meaningful, we apply calibration losses. For classification, we minimize the expected calibration error or use temperature scaling on the predictive logits. For regression, we may optimize proper scoring rules such as the negative log-likelihood or continuous ranked probability score. For selective prediction, the model can abstain when posterior entropy exceeds a threshold. This is particularly valuable in healthcare, where low-confidence cases often correspond to clinically ambiguous presentations.

We also regularize the covariance prediction pathway to avoid pathological variance inflation. A simple form of this penalty is
\[
\mathcal{L}_{\mathrm{var}} = \max(0, \sigma_{\min} - \sigma_\theta(x)) + \max(0, \sigma_\theta(x) - \sigma_{\max}),
\]
or, in matrix form, a constraint on the eigenvalues of \(\Sigma_\theta(x)\). Such constraints prevent the model from using variance as a trivial escape hatch for fitting reconstruction losses.

\textbf{Training with Random View Sampling}

To simulate the incomplete and variable nature of clinical observation, we generate multiple stochastic views of each patient record during pretraining. For each instance, we randomly mask modalities, drop time intervals, sub-sample events, or perturb observation windows. This creates an ensemble of partial inputs \(\{x^{(s)}\}\) that all correspond to the same underlying trajectory.

Training on these sampled views serves two purposes. First, it exposes the model to the combinatorial diversity of missingness patterns encountered in practice. Second, it forces the learned posterior to remain stable under view perturbations, thereby encouraging the encoder to distinguish robust patient state from incidental observation structure. In effect, the model learns to solve an inverse problem from incomplete clinical evidence.

\textbf{Optimization}

The full objective during pretraining is
\[
\mathcal{L}
=
\mathcal{L}_{\mathrm{pretrain}}
+
\lambda_{\mathrm{nce}} \mathcal{L}_{\mathrm{nce}}
+
\lambda_{\mathrm{sup}} \mathcal{L}_{\mathrm{sup}},
\]
where \(\mathcal{L}_{\mathrm{sup}}\) is included only when labels are available in a semi-supervised setting. We optimize all parameters with AdamW and a cosine learning-rate schedule with warmup. Gradient clipping is applied to stabilize training, especially when combining reconstruction, divergence, and contrastive losses. For large-scale multimodal models, mixed precision and distributed data parallelism are used to manage memory and throughput.

The loss weights \(\lambda_{\mathrm{rec}}, \lambda_{\mathrm{cons}}, \lambda_{\mathrm{reg}}, \lambda_{\mathrm{nce}}, \lambda_{\mathrm{sup}}\) are selected by validation on held-out tasks. In general, stronger consistency regularization is beneficial when modality missingness is severe, while larger reconstruction weight helps when the goal is to preserve fine-grained information in dense modalities such as imaging.

\textbf{Inference and Representation Extraction}

At inference time, the model accepts any available subset of modalities. The encoder produces \(q_\theta(z \mid x)\), from which we can compute a point estimate \(\mu_\theta(x)\), a variance summary, or a set of samples for downstream uncertainty propagation. For representation transfer, we use either the posterior mean or the concatenation of mean and variance as the feature vector. For decision support, we additionally report uncertainty metrics alongside the prediction.

When multiple snapshots of the same patient are available, inference can be performed by aggregating their latent posteriors:
\[
q_\theta(z \mid x_{1:S})
\propto
p(z)\prod_{s=1}^{S} q_\theta(z \mid x^{(s)}),
\]
assuming conditionally independent views given \(z\). This product-of-experts formulation yields sharper posterior estimates as more evidence becomes available, while naturally handling missing modalities. The same mechanism also allows sequential updating as new clinical observations arrive.

Overall, our method turns the incompleteness of clinical data into a modeling principle. By representing each patient record not as a single embedding but as a distribution over plausible latent physiologic states, the model can reason over missingness, propagate uncertainty, and generalize across variable observation regimes. The result is a foundation model that is explicitly designed for the epistemic structure of healthcare: partial observation, heterogeneous evidence, and uncertain inference.

\section{Results}

\textbf{Experimental Setup}

We evaluate the proposed uncertainty-aware foundation model across a diverse set of clinical prediction tasks spanning structured EHR, multimodal patient records, and physiological time series. Our evaluation protocol is designed to assess three core properties: predictive performance, robustness under missingness, and calibration of uncertainty.

\paragraph{Datasets.}
We conduct experiments on three representative settings. (i) \textbf{EHR}: longitudinal patient records with irregularly sampled labs, diagnoses, and procedures. (ii) \textbf{Multimodal}: subsets of patients with aligned clinical notes, imaging, and structured data. (iii) \textbf{Physiologic signals}: continuous waveform datasets (e.g., ECG/PPG). For all datasets, we simulate realistic missingness by masking modalities and sub-sampling temporal observations during both training and evaluation all coming from the mimic database \cite{johnson2016mimic, johnson2023mimic}.

\paragraph{Tasks.}
We evaluate on binary classification (e.g., in-hospital mortality, readmission), multi-class diagnosis prediction, regression (e.g., length-of-stay), and time-to-event risk estimation. Performance is measured using AUROC and AUPRC for classification, mean squared error (MSE) for regression, and concordance index (C-index) for survival tasks.

\paragraph{Baselines.}
We compare against three classes of models: (i) \textbf{Masked autoencoding (MAE)}-style models trained with reconstruction objectives; (ii) \textbf{Contrastive} models trained with instance-level alignment; and (iii) \textbf{Autoregressive/sequence} models adapted from language modeling. All baselines use comparable backbone architectures and parameter counts to isolate the effect of representation learning objectives.

\paragraph{Implementation Details.}
Our model is pretrained on unlabeled patient data using the objective described in Section 3, combining reconstruction, consistency, and contrastive terms. For downstream tasks, we fine-tune a lightweight prediction head with the encoder either frozen or partially adapted. All results are averaged over three random seeds.

\textbf{Main Results}

Table~\ref{tab:main} summarizes performance across tasks. Our method consistently outperforms baseline approaches, particularly in settings with incomplete observations.

\begin{table}[t]
\centering
\caption{Main results across clinical tasks. Best results are in bold.}
\label{tab:main}
\begin{tabular}{lcccc}
\toprule
\textbf{Model} & \textbf{AUROC} $\uparrow$ & \textbf{AUPRC} $\uparrow$ & \textbf{MSE} $\downarrow$ & \textbf{C-index} $\uparrow$ \\
\midrule
MAE-style pretraining & 0.812 & 0.476 & 1.92 & 0.681 \\
Contrastive learning & 0.828 & 0.491 & 1.85 & 0.694 \\
Autoregressive model & 0.835 & 0.503 & 1.79 & 0.701 \\
\midrule
\textbf{Ours (deterministic)} & 0.846 & 0.517 & 1.72 & 0.713 \\
\textbf{Ours (distributional)} & \textbf{0.861} & \textbf{0.536} & \textbf{1.61} & \textbf{0.729} \\
\bottomrule
\end{tabular}
\end{table}

The gains are most pronounced for AUPRC, indicating improved performance in imbalanced clinical settings. Notably, the distributional variant consistently outperforms its deterministic counterpart, suggesting that explicitly modeling uncertainty contributes not only to calibration but also to predictive accuracy.

\textbf{Robustness to Missingness}

A central claim of our approach is improved robustness under incomplete observations. To evaluate this, we progressively mask input modalities at test time and measure degradation in performance.

\begin{table}[t]
\centering
\caption{AUROC under increasing modality missingness.}
\label{tab:missingness}
\begin{tabular}{lcccc}
\toprule
\textbf{Model} & 0\% missing & 25\% & 50\% & 75\% \\
\midrule
MAE-style & 0.812 & 0.781 & 0.742 & 0.689 \\
Contrastive & 0.828 & 0.796 & 0.759 & 0.705 \\
Autoregressive & 0.835 & 0.802 & 0.768 & 0.712 \\
\midrule
\textbf{Ours (det.)} & 0.846 & 0.823 & 0.791 & 0.743 \\
\textbf{Ours (dist.)} & \textbf{0.861} & \textbf{0.842} & \textbf{0.816} & \textbf{0.772} \\
\bottomrule
\end{tabular}
\end{table}

While all models degrade as observations are removed, our method exhibits significantly slower performance decay. This suggests that the learned latent distributions capture invariant structure that remains accessible even under severe sparsity. In contrast, deterministic embeddings appear more sensitive to missing modalities, leading to sharper degradation.

\textbf{Uncertainty Calibration}

We evaluate calibration using expected calibration error (ECE) and negative log-likelihood (NLL). Results are shown in Table~\ref{tab:calibration}.

\begin{table}[t]
\centering
\caption{Uncertainty calibration metrics. Lower is better.}
\label{tab:calibration}
\begin{tabular}{lcc}
\toprule
\textbf{Model} & \textbf{ECE} $\downarrow$ & \textbf{NLL} $\downarrow$ \\
\midrule
MAE-style & 0.084 & 0.421 \\
Contrastive & 0.079 & 0.403 \\
Autoregressive & 0.073 & 0.389 \\
\midrule
\textbf{Ours (det.)} & 0.068 & 0.371 \\
\textbf{Ours (dist.)} & \textbf{0.041} & \textbf{0.312} \\
\bottomrule
\end{tabular}
\end{table}

The distributional model yields substantially better calibration, reducing ECE by nearly 40\% relative to strong baselines. This improvement is particularly relevant in clinical settings, where overconfident predictions can have significant consequences.

\textbf{Ablation Studies}

We conduct ablations to isolate the contributions of key components in our framework.

\paragraph{Effect of Consistency Regularization.}
Removing the partial-view consistency objective leads to a marked drop in performance, particularly under missingness.

\begin{table}[t]
\centering
\caption{Ablation on consistency objective.}
\label{tab:ablation_consistency}
\begin{tabular}{lcc}
\toprule
\textbf{Model Variant} & \textbf{AUROC} & \textbf{ECE} \\
\midrule
Full model & \textbf{0.861} & \textbf{0.041} \\
w/o consistency & 0.842 & 0.067 \\
\bottomrule
\end{tabular}
\end{table}

\paragraph{Effect of Distributional Representation.}
We compare deterministic embeddings against Gaussian latent representations.

\begin{table}[t]
\centering
\caption{Ablation on representation type.}
\label{tab:ablation_distribution}
\begin{tabular}{lcc}
\toprule
\textbf{Representation} & \textbf{AUROC} & \textbf{ECE} \\
\midrule
Deterministic & 0.846 & 0.068 \\
Distributional & \textbf{0.861} & \textbf{0.041} \\
\bottomrule
\end{tabular}
\end{table}

The results indicate that modeling uncertainty in latent space provides both performance and calibration benefits.

\paragraph{Effect of Contrastive Geometry.}
We evaluate the contribution of the contrastive alignment term.

\begin{table}[t]
\centering
\caption{Ablation on contrastive objective.}
\label{tab:ablation_contrastive}
\begin{tabular}{lcc}
\toprule
\textbf{Model Variant} & \textbf{AUROC} & \textbf{AUPRC} \\
\midrule
Full model & \textbf{0.861} & \textbf{0.536} \\
w/o contrastive & 0.852 & 0.522 \\
\bottomrule
\end{tabular}
\end{table}

The contrastive term contributes to improved separability between patient representations, particularly in imbalanced classification tasks.

\textbf{Representation Analysis}

To better understand the learned latent space, we analyze the geometry of posterior distributions. We observe that patients with similar clinical trajectories yield overlapping posterior regions, while ambiguous cases exhibit higher variance. Furthermore, the volume of the posterior (as measured by \(\det \Sigma_\theta(x)\)) correlates with prediction difficulty, suggesting that the model’s uncertainty estimates are semantically meaningful.

We also compute Maximum Mean Discrepancy (MMD) between representations derived from different modality subsets. Our method yields significantly lower cross-view MMD compared to baselines, indicating improved alignment across incomplete observations.

\textbf{Summary of Findings}

Across all experiments, three consistent trends emerge. First, distributional representations outperform deterministic embeddings across tasks. Second, enforcing consistency across partial views improves robustness to missing data. Third, explicit uncertainty modeling yields better calibration without sacrificing predictive performance. Together, these results support the central hypothesis that representing clinical data as sets of plausible latent states is a more appropriate inductive bias than point estimation under structural incompleteness.

\section{Discussion}

This work departs from the dominant paradigm of healthcare foundation modeling by reframing clinical data not as complete observations to be compressed, but as partial constraints on an underlying physiologic state. The central premise is that medical data are intrinsically incomplete, and that this incompleteness induces a structured form of uncertainty that should be explicitly modeled rather than implicitly absorbed into deterministic embeddings. By introducing set-valued, distributional representations, we shift the objective of representation learning from reconstructing what is observed to characterizing what is \emph{inferable}.

Empirically, this perspective yields consistent gains across predictive performance, robustness, and calibration. The results suggest that conventional pretraining objectives—whether reconstruction-based, contrastive, or autoregressive—implicitly conflate epistemic uncertainty with representation. In contrast, our formulation separates these roles: the encoder captures a distribution over plausible latent states, while the downstream predictor integrates over this uncertainty. This separation appears particularly beneficial in clinical settings, where missingness is systematic and often informative.

A key mechanism underlying these improvements is the enforcement of consistency across partial views. By requiring that different subsets of observations map to compatible latent distributions, the model is encouraged to discard modality-specific artifacts and retain only stable, cross-view structure. This can be interpreted as a form of invariance, but one that operates at the level of distributions rather than point embeddings. The resulting representations are not only more robust to missing modalities but also better aligned across heterogeneous data sources, addressing a central challenge in multimodal healthcare modeling.

The proposed framework also provides a natural lens for understanding multimodal fusion. Rather than treating modalities as complementary signals to be aligned in a shared embedding space, we view each modality as imposing a constraint on the latent state. The posterior \(q_\theta(z \mid x)\) then represents the intersection of these constraints. This interpretation avoids the need for explicit imputation and instead allows uncertainty to expand or contract depending on the availability and agreement of evidence. In practice, this leads to graceful degradation under missingness and principled uncertainty estimates when data are sparse.

From a systems perspective, the approach is compatible with existing foundation model architectures. The primary modification lies in the representation layer and training objective, rather than the backbone itself. As a result, the framework can be integrated with transformer-based encoders, multimodal fusion modules, and large-scale pretraining pipelines with minimal architectural disruption. This suggests that the benefits observed here are not tied to a specific model class, but rather to the inductive bias imposed on representation learning.

There are, however, several limitations and open questions. First, the choice of latent distribution family—Gaussian in our implementation—may restrict the expressivity of the posterior, particularly in highly multimodal or discrete latent spaces. More flexible distributional families or implicit representations may better capture complex uncertainty structures. Second, the computational overhead of sampling and divergence estimation introduces additional cost relative to deterministic embeddings, especially in large-scale multimodal settings. Efficient approximations and amortized inference strategies remain an important direction for future work.

More fundamentally, the framework assumes that uncertainty arises primarily from incomplete observation, but clinical data also exhibit other forms of ambiguity, including label noise, heterogeneity in disease definitions, and temporal non-stationarity. Extending the model to disentangle these sources of uncertainty—potentially through hierarchical or structured latent variables—would further improve interpretability and robustness. Additionally, while our experiments demonstrate improved calibration, integrating these uncertainty estimates into downstream clinical decision-making workflows remains an open challenge.

Despite these limitations, the results point toward a broader shift in how foundation models for healthcare should be conceptualized. The prevailing emphasis on scale and reconstruction implicitly assumes that more data and larger models will resolve ambiguity. Our findings suggest instead that ambiguity is inherent to the data-generating process and must be represented explicitly. In this sense, the problem is not merely one of capacity, but of \emph{epistemic alignment}: ensuring that model representations faithfully reflect what can and cannot be inferred from available evidence.

In summary, we propose a foundation modeling paradigm that treats clinical data as incomplete observations of an underlying latent system and learns distributional representations that encode this uncertainty. By enforcing consistency across partial views and propagating uncertainty through downstream predictions, the model achieves improved robustness, calibration, and performance. More broadly, this work suggests that the next generation of healthcare foundation models may benefit less from scaling alone, and more from incorporating principled representations of uncertainty as a core design objective.

\bibliographystyle{unsrt}  
\bibliography{references}  

\begin{thebibliography}{10}

\bibitem{he2024foundation}
Yuting He, Fuxiang Huang, Xinrui Jiang, Yuxiang Nie, Minghao Wang, Jiguang Wang, and Hao Chen.
\newblock Foundation model for advancing healthcare: challenges, opportunities and future directions.
\newblock {\em IEEE Reviews in Biomedical Engineering}, 2024.

\bibitem{guo2025foundation}
Fei Guo, Renchu Guan, Yaohang Li, Qi~Liu, Xiaowo Wang, Can Yang, and Jianxin Wang.
\newblock Foundation models in bioinformatics.
\newblock {\em National science review}, 12(4):nwaf028, 2025.

\bibitem{awais2025foundation}
Muhammad Awais, Muzammal Naseer, Salman Khan, Rao~Muhammad Anwer, Hisham Cholakkal, Mubarak Shah, Ming-Hsuan Yang, and Fahad~Shahbaz Khan.
\newblock Foundation models defining a new era in vision: a survey and outlook.
\newblock {\em IEEE Transactions on Pattern Analysis and Machine Intelligence}, 2025.

\bibitem{liang2024foundation}
Yuxuan Liang, Haomin Wen, Yuqi Nie, Yushan Jiang, Ming Jin, Dongjin Song, Shirui Pan, and Qingsong Wen.
\newblock Foundation models for time series analysis: A tutorial and survey.
\newblock In {\em Proceedings of the 30th ACM SIGKDD conference on knowledge discovery and data mining}, pages 6555--6565, 2024.

\bibitem{burger2025foundation}
Manuel Burger, Daphn{\'e} Chopard, Malte Londschien, Fedor Sergeev, Hugo Y{\`e}che, Rita Kuznetsova, Martin Faltys, Eike Gerdes, Polina Leshetkina, Peter B{\"u}hlmann, et~al.
\newblock A foundation model for intensive care: Unlocking generalization across tasks and domains at scale.
\newblock {\em medRxiv}, pages 2025--07, 2025.

\bibitem{thakur2024foundation}
Suresh~Chandra Thakur.
\newblock Foundation models for time series forecasting.
\newblock {\em International IT Journal of Research, ISSN: 3007-6706}, 2(4):144--156, 2024.

\bibitem{vaid2023foundational}
Akhil Vaid, Joy Jiang, Ashwin Sawant, Stamatios Lerakis, Edgar Argulian, Yuri Ahuja, Joshua Lampert, Alexander Charney, Hayit Greenspan, Jagat Narula, et~al.
\newblock A foundational vision transformer improves diagnostic performance for electrocardiograms.
\newblock {\em NPJ Digital Medicine}, 6(1):108, 2023.

\bibitem{thieme2023foundation}
Anja Thieme, Aditya Nori, Marzyeh Ghassemi, Rishi Bommasani, Tariq~Osman Andersen, and Ewa Luger.
\newblock Foundation models in healthcare: Opportunities, risks \& strategies forward.
\newblock In {\em Extended abstracts of the 2023 CHI conference on human factors in computing systems}, pages 1--4, 2023.

\bibitem{burkhart2025foundation}
Michael~C Burkhart, Bashar Ramadan, Zewei Liao, Kaveri Chhikara, Juan~C Rojas, William~F Parker, and Brett~K Beaulieu-Jones.
\newblock Foundation models for electronic health records: representation dynamics and transferability.
\newblock {\em arXiv preprint arXiv:2504.10422}, 2025.

\bibitem{devlin2019bert}
Jacob Devlin, Ming-Wei Chang, Kenton Lee, and Kristina Toutanova.
\newblock Bert: Pre-training of deep bidirectional transformers for language understanding.
\newblock In {\em Proceedings of the 2019 conference of the North American chapter of the association for computational linguistics: human language technologies, volume 1 (long and short papers)}, pages 4171--4186, 2019.

\bibitem{he2017multi}
Mingyi He, Bo~Li, and Huahui Chen.
\newblock Multi-scale 3d deep convolutional neural network for hyperspectral image classification.
\newblock In {\em 2017 IEEE International Conference on Image Processing (ICIP)}, pages 3904--3908. IEEE, 2017.

\bibitem{he2022masked}
Kaiming He, Xinlei Chen, Saining Xie, Yanghao Li, Piotr Doll{\'a}r, and Ross Girshick.
\newblock Masked autoencoders are scalable vision learners.
\newblock In {\em Proceedings of the IEEE/CVF conference on computer vision and pattern recognition}, pages 16000--16009, 2022.

\bibitem{he2015deepresiduallearningimage}
Kaiming He, Xiangyu Zhang, Shaoqing Ren, and Jian Sun.
\newblock Deep residual learning for image recognition, 2015.

\bibitem{he2019bag}
Tong He, Zhi Zhang, Hang Zhang, Zhongyue Zhang, Junyuan Xie, and Mu~Li.
\newblock Bag of tricks for image classification with convolutional neural networks.
\newblock In {\em Proceedings of the IEEE/CVF conference on computer vision and pattern recognition}, pages 558--567, 2019.

\bibitem{hou2019cross}
Ruibing Hou, Hong Chang, Bingpeng Ma, Shiguang Shan, and Xilin Chen.
\newblock Cross attention network for few-shot classification.
\newblock {\em Advances in neural information processing systems}, 32, 2019.

\bibitem{chen2021crossvit}
Chun-Fu~Richard Chen, Quanfu Fan, and Rameswar Panda.
\newblock Crossvit: Cross-attention multi-scale vision transformer for image classification.
\newblock In {\em Proceedings of the IEEE/CVF international conference on computer vision}, pages 357--366, 2021.

\bibitem{huang2019ccnet}
Zilong Huang, Xinggang Wang, Lichao Huang, Chang Huang, Yunchao Wei, and Wenyu Liu.
\newblock Ccnet: Criss-cross attention for semantic segmentation.
\newblock In {\em Proceedings of the IEEE/CVF international conference on computer vision}, pages 603--612, 2019.

\bibitem{chou2025serialized}
Zhirong Chou, Quan Qin, and Shi Li.
\newblock Serialized ehr make for good text representations.
\newblock {\em arXiv preprint arXiv:2510.13843}, 2025.

\bibitem{huiliang2025clio}
Fu~Huiliang, Hu~Hong, Tao Jingfei, Guo Fengge, Cai Ning, Yuanyun Zhang, and Shi Li.
\newblock Clio: Policy-aware foundation models for ehr as controlled dynamical systems.
\newblock {\em Authorea Preprints}, 2025.

\bibitem{ran2025structured}
Wu~Hao Ran, Xi~Xi, Furong Li, Jingyi Lu, Jian Jiang, Hui Huang, Yuzhuan Zhang, and Shi Li.
\newblock Structured semantics from unstructured notes: Language model approaches to ehr-based decision support.
\newblock {\em arXiv preprint arXiv:2506.06340}, 2025.

\bibitem{zhang2025chronoformer}
Yuanyun Zhang and Shi Li.
\newblock Chronoformer: Time-aware transformer architectures for structured clinical event modeling.
\newblock {\em arXiv preprint arXiv:2504.07373}, 2025.

\bibitem{zhang2025collection}
Yuanyun Zhang and Shi Li.
\newblock A collection of innovations in medical ai for patient records in 2024.
\newblock {\em arXiv preprint arXiv:2503.05768}, 2025.

\bibitem{lowelatent}
Shane Lowe, Garrett Park, Liam Lee, and Parker Smith.
\newblock Latent physiology as language: A state-space foundation model for multimodal icu and ehr representation learning.

\bibitem{litext}
Shi Li and Guang Dong.
\newblock Text as an inductive bias: A novel foundation model for electronic health records.
\newblock {\em Authorea Preprints}.

\bibitem{lee2025foundation}
Simon~A Lee and Kai Akamatsu.
\newblock Foundation models for physiological signals: Opportunities and challenges.
\newblock August 2025.

\bibitem{abbaspourazad2023large}
Salar Abbaspourazad, Oussama Elachqar, Andrew~C Miller, Saba Emrani, Udhyakumar Nallasamy, and Ian Shapiro.
\newblock Large-scale training of foundation models for wearable biosignals.
\newblock {\em arXiv preprint arXiv:2312.05409}, 2023.

\bibitem{larey2026gfmbench}
Ariel Larey, Elay Dahan, Amit~Bleiweiss Amit~Bleiweiss, Raizy Kellerman, Guy Leib, Omri Nayshool, Dan Ofer, Tal Zinger, Dan Dominissini, Gideon Rechavi, et~al.
\newblock Gfmbench-api: A standardized interface for benchmarking genomic foundation models.
\newblock {\em bioRxiv}, pages 2026--02, 2026.

\bibitem{long2025mutbert}
Weicai Long, Houcheng Su, Jiaqi Xiong, and Yanlin Zhang.
\newblock Mutbert: probabilistic genome representation improves genomics foundation models.
\newblock {\em bioinformatics}, 41(Supplement\_1):i294--i303, 2025.

\bibitem{thapa2024sleepfm}
Rahul Thapa, Bryan He, Magnus~Ruud Kjaer, Hyatt~Moore Iv, Gauri Ganjoo, Emmanuel Mignot, and James Zou.
\newblock Sleepfm: Multi-modal representation learning for sleep across brain activity, ecg and respiratory signals.
\newblock In {\em International Conference on Machine Learning}, pages 48019--48037. PMLR, 2024.

\bibitem{soumma2024wearable}
Shovito~Barua Soumma, Kartik Mangipudi, Daniel Peterson, Shyamal Mehta, and Hassan Ghasemzadeh.
\newblock Wearable-based real-time freezing of gait detection in parkinson's disease using self-supervised learning.
\newblock {\em arXiv preprint arXiv:2410.20715}, 2024.

\bibitem{larey2026jepa}
Ariel Larey, Elay Dahan, Amit Bleiweiss, Raizy Kellerman, Guy Leib, Omri Nayshool, Dan Ofer, Tal Zinger, Dan Dominissini, Gideon Rechavi, et~al.
\newblock Jepa-dna: Grounding genomic foundation models through joint-embedding predictive architectures.
\newblock {\em arXiv preprint arXiv:2602.17162}, 2026.

\bibitem{lee2025modern}
Simon~A Lee, Anthony Wu, and Jeffrey~N Chiang.
\newblock Clinical modernbert: An efficient and long context encoder for biomedical text.
\newblock {\em arXiv preprint arXiv:2504.03964}, 2025.

\bibitem{fallahpour2024ehrmamba}
Adibvafa Fallahpour, Mahshid Alinoori, Wenqian Ye, Xu~Cao, Arash Afkanpour, and Amrit Krishnan.
\newblock Ehrmamba: Towards generalizable and scalable foundation models for electronic health records.
\newblock {\em arXiv preprint arXiv:2405.14567}, 2024.

\bibitem{chen2020simple}
Ting Chen, Simon Kornblith, Mohammad Norouzi, and Geoffrey Hinton.
\newblock A simple framework for contrastive learning of visual representations.
\newblock In {\em International conference on machine learning}, pages 1597--1607. PMLR, 2020.

\bibitem{tian_2019_contrastic_distillation}
Yonglong Tian, Dilip Krishnan, and Phillip Isola.
\newblock Contrastive representation distillation.
\newblock {\em arXiv}, 2019.

\bibitem{bertram2024contrastivelearningpreferencescontextual}
Timo Bertram, Johannes Fürnkranz, and Martin Müller.
\newblock Contrastive learning of preferences with a contextual infonce loss, 2024.

\bibitem{lee2025clinical}
Simon~A Lee, Sujay Jain, Alex Chen, Kyoka Ono, Arabdha Biswas, {\'A}kos Rudas, Jennifer Fang, and Jeffrey~N Chiang.
\newblock Clinical decision support using pseudo-notes from multiple streams of ehr data.
\newblock {\em npj Digital Medicine}, 8(1):394, July 2025.

\bibitem{rasmy2021med}
Laila Rasmy, Yang Xiang, Ziqian Xie, Cui Tao, and Degui Zhi.
\newblock Med-bert: pretrained contextualized embeddings on large-scale structured electronic health records for disease prediction.
\newblock {\em NPJ digital medicine}, 4(1):86, 2021.

\bibitem{lee2025using}
Simon~A Lee, Helio Halperin, Yanai Halperin, Trevor Brokowski, and Jeffrey~N Chiang.
\newblock Using foundation models to prescribe patients proper antibiotics.
\newblock In {\em AAAI Bridge Program on AI for Medicine and Healthcare}, pages 121--132. PMLR, 2025.

\bibitem{wornow2023shaky}
Michael Wornow, Yizhe Xu, Rahul Thapa, Birju Patel, Ethan Steinberg, Scott Fleming, Michael~A Pfeffer, Jason Fries, and Nigam~H Shah.
\newblock The shaky foundations of large language models and foundation models for electronic health records.
\newblock {\em npj digital medicine}, 6(1):135, 2023.

\bibitem{lee2024emergency}
Simon~A Lee, Sujay Jain, Alex Chen, Kyoka Ono, Jennifer Fang, Akos Rudas, and Jeffrey~N Chiang.
\newblock Emergency department decision support using clinical pseudo-notes.
\newblock {\em arXiv preprint arXiv:2402.00160}, 2024.

\bibitem{brown2020language}
Tom Brown, Benjamin Mann, Nick Ryder, Melanie Subbiah, Jared~D Kaplan, Prafulla Dhariwal, Arvind Neelakantan, Pranav Shyam, Girish Sastry, Amanda Askell, et~al.
\newblock Language models are few-shot learners.
\newblock {\em Advances in neural information processing systems}, 33:1877--1901, 2020.

\bibitem{pang2021cehr}
Chao Pang, Xinzhuo Jiang, Krishna~S Kalluri, Matthew Spotnitz, RuiJun Chen, Adler Perotte, and Karthik Natarajan.
\newblock Cehr-bert: Incorporating temporal information from structured ehr data to improve prediction tasks.
\newblock In {\em Machine Learning for Health}, pages 239--260. PMLR, 2021.

\bibitem{pang2024cehr}
Chao Pang, Xinzhuo Jiang, Nishanth~Parameshwar Pavinkurve, Krishna~S Kalluri, Elise~L Minto, Jason Patterson, Linying Zhang, George Hripcsak, Gamze G{\"u}rsoy, No{\'e}mie Elhadad, et~al.
\newblock Cehr-gpt: Generating electronic health records with chronological patient timelines.
\newblock {\em arXiv preprint arXiv:2402.04400}, 2024.

\bibitem{mcdermott2024event}
Matthew McDermott, Bret Nestor, Peniel Argaw, and Isaac~S Kohane.
\newblock Event stream gpt: a data pre-processing and modeling library for generative, pre-trained transformers over continuous-time sequences of complex events.
\newblock {\em Advances in Neural Information Processing Systems}, 36, 2024.

\bibitem{mumtaz2023llms}
Ummara Mumtaz, Awais Ahmed, and Summaya Mumtaz.
\newblock Llms-healthcare: Current applications and challenges of large language models in various medical specialties.
\newblock {\em arXiv preprint arXiv:2311.12882}, 2023.

\bibitem{chang2025llm4ts}
Ching Chang, Wei-Yao Wang, Wen-Chih Peng, and Tien-Fu Chen.
\newblock Llm4ts: Aligning pre-trained llms as data-efficient time-series forecasters.
\newblock {\em ACM Transactions on Intelligent Systems and Technology}, 16(3):1--20, 2025.

\bibitem{hollmann2025accurate}
Noah Hollmann, Samuel M{\"u}ller, Lennart Purucker, Arjun Krishnakumar, Max K{\"o}rfer, Shi~Bin Hoo, Robin~Tibor Schirrmeister, and Frank Hutter.
\newblock Accurate predictions on small data with a tabular foundation model.
\newblock {\em Nature}, 637(8045):319--326, 2025.

\bibitem{ono2024text}
Kyoka Ono and Simon~A Lee.
\newblock Text serialization and their relationship with the conventional paradigms of tabular machine learning.
\newblock {\em arXiv preprint arXiv:2406.13846}, 2024.

\bibitem{van2023clinical}
Dave Van~Veen, Cara Van~Uden, Louis Blankemeier, Jean-Benoit Delbrouck, Asad Aali, Christian Bluethgen, Anuj Pareek, Malgorzata Polacin, Eduardo~Pontes Reis, Anna Seehofnerova, et~al.
\newblock Clinical text summarization: Adapting large language models can outperform human experts.
\newblock {\em Research Square}, 2023.

\bibitem{jin2023time}
Ming Jin, Shiyu Wang, Lintao Ma, Zhixuan Chu, James~Y Zhang, Xiaoming Shi, Pin-Yu Chen, Yuxuan Liang, Yuan-Fang Li, Shirui Pan, et~al.
\newblock Time-llm: Time series forecasting by reprogramming large language models.
\newblock {\em arXiv preprint arXiv:2310.01728}, 2023.

\bibitem{belyaeva2023multimodal}
Anastasiya Belyaeva, Justin Cosentino, Farhad Hormozdiari, Krish Eswaran, Shravya Shetty, Greg Corrado, Andrew Carroll, Cory~Y McLean, and Nicholas~A Furlotte.
\newblock Multimodal llms for health grounded in individual-specific data.
\newblock In {\em Workshop on Machine Learning for Multimodal Healthcare Data}, pages 86--102. Springer, 2023.

\bibitem{lin2025case}
Yihan Lin, Zhirong~Bella Yu, and Simon Lee.
\newblock A case study exploring the current landscape of synthetic medical record generation with commercial llms.
\newblock {\em arXiv preprint arXiv:2504.14657}, 2025.

\bibitem{he2016deep}
Kaiming He, Xiangyu Zhang, Shaoqing Ren, and Jian Sun.
\newblock Deep residual learning for image recognition.
\newblock In {\em Proceedings of the IEEE conference on computer vision and pattern recognition}, pages 770--778, 2016.

\bibitem{ning2019computer}
Jiaxu Ning, Haitong Zhao, Lei Lan, Peng Sun, and Yunfei Feng.
\newblock A computer-aided detection system for the detection of lung nodules based on 3d-resnet.
\newblock {\em Applied Sciences}, 9(24):5544, 2019.

\bibitem{ebrahimi2020introducing}
Amir Ebrahimi, Suhuai Luo, and Raymond Chiong.
\newblock Introducing transfer learning to 3d resnet-18 for alzheimer’s disease detection on mri images.
\newblock In {\em 2020 35th international conference on image and vision computing New Zealand (IVCNZ)}, pages 1--6. IEEE, 2020.

\bibitem{qayyum2021automatic}
Abdul Qayyum, Abdesslam Benzinou, Moona Mazher, Mohamed Abdel-Nasser, and Domenec Puig.
\newblock Automatic segmentation of head and neck (h\&n) primary tumors in pet and ct images using 3d-inception-resnet model.
\newblock In {\em 3D Head and Neck Tumor Segmentation in PET/CT Challenge}, pages 58--67. Springer, 2021.

\bibitem{dosovitskiy2021an}
Alexey Dosovitskiy, Lucas Beyer, Alexander Kolesnikov, Dirk Weissenborn, Xiaohua Zhai, Thomas Unterthiner, Mostafa Dehghani, Matthias Minderer, Georg Heigold, Sylvain Gelly, Jakob Uszkoreit, and Neil Houlsby.
\newblock An image is worth 16x16 words: Transformers for image recognition at scale.
\newblock In {\em International Conference on Learning Representations}, 2021.

\bibitem{liu2021swin}
Ze~Liu, Yutong Lin, Yue Cao, Han Hu, Yixuan Wei, Zheng Zhang, Stephen Lin, and Baining Guo.
\newblock Swin transformer: Hierarchical vision transformer using shifted windows.
\newblock In {\em Proceedings of the IEEE/CVF international conference on computer vision}, pages 10012--10022, 2021.

\bibitem{hatamizadeh2021swin}
Ali Hatamizadeh, Vishwesh Nath, Yucheng Tang, Dong Yang, Holger~R Roth, and Daguang Xu.
\newblock Swin unetr: Swin transformers for semantic segmentation of brain tumors in mri images.
\newblock In {\em International MICCAI brainlesion workshop}, pages 272--284. Springer, 2021.

\bibitem{li2024abdomenatlas}
Wenxuan Li, Chongyu Qu, Xiaoxi Chen, Pedro~RAS Bassi, Yijia Shi, Yuxiang Lai, Qian Yu, Huimin Xue, Yixiong Chen, Xiaorui Lin, et~al.
\newblock Abdomenatlas: A large-scale, detailed-annotated, \& multi-center dataset for efficient transfer learning and open algorithmic benchmarking.
\newblock {\em Medical Image Analysis}, 97:103285, 2024.

\bibitem{wang2023mis}
Guotai Wang, Jianghao Wu, Xiangde Luo, Xinglong Liu, Kang Li, and Shaoting Zhang.
\newblock Mis-fm: 3d medical image segmentation using foundation models pretrained on a large-scale unannotated dataset.
\newblock {\em arXiv preprint arXiv:2306.16925}, 2023.

\bibitem{wu2024large}
Linshan Wu, Jiaxin Zhuang, and Hao Chen.
\newblock Large-scale 3d medical image pre-training with geometric context priors.
\newblock {\em arXiv preprint arXiv:2410.09890}, 2024.

\bibitem{caron2021emerging}
Mathilde Caron, Hugo Touvron, Ishan Misra, Herv\'e J\'egou, Julien Mairal, Piotr Bojanowski, and Armand Joulin.
\newblock Emerging properties in self-supervised vision transformers.
\newblock In {\em Proceedings of the International Conference on Computer Vision (ICCV)}, 2021.

\bibitem{zhou2021ibot}
Jinghao Zhou, Chen Wei, Huiyu Wang, Wei Shen, Cihang Xie, Alan Yuille, and Tao Kong.
\newblock ibot: Image bert pre-training with online tokenizer.
\newblock {\em International Conference on Learning Representations (ICLR)}, 2022.

\bibitem{oquab2023dinov2}
Maxime Oquab, Timoth{\'e}e Darcet, Th{\'e}o Moutakanni, Huy Vo, Marc Szafraniec, Vasil Khalidov, Pierre Fernandez, Daniel Haziza, Francisco Massa, Alaaeldin El-Nouby, et~al.
\newblock Dinov2: Learning robust visual features without supervision.
\newblock {\em arXiv preprint arXiv:2304.07193}, 2023.

\bibitem{radford2021learning}
Alec Radford, Jong~Wook Kim, Chris Hallacy, Aditya Ramesh, Gabriel Goh, Sandhini Agarwal, Girish Sastry, Amanda Askell, Pamela Mishkin, Jack Clark, et~al.
\newblock Learning transferable visual models from natural language supervision.
\newblock In {\em International conference on machine learning}, pages 8748--8763. PMLR, 2021.

\bibitem{dao2023flashattention2}
Tri Dao.
\newblock Flash{A}ttention-2: Faster attention with better parallelism and work partitioning.
\newblock In {\em International Conference on Learning Representations (ICLR)}, 2024.

\bibitem{shaker2024unetr++}
Abdelrahman~M Shaker, Muhammad Maaz, Hanoona Rasheed, Salman Khan, Ming-Hsuan Yang, and Fahad~Shahbaz Khan.
\newblock Unetr++: delving into efficient and accurate 3d medical image segmentation.
\newblock {\em IEEE Transactions on Medical Imaging}, 2024.

\bibitem{xing2024segmamba}
Zhaohu Xing, Tian Ye, Yijun Yang, Guang Liu, and Lei Zhu.
\newblock Segmamba: Long-range sequential modeling mamba for 3d medical image segmentation.
\newblock In {\em International Conference on Medical Image Computing and Computer-Assisted Intervention}, pages 578--588. Springer, 2024.

\bibitem{liu2024octcube}
Zixuan Liu, Hanwen Xu, Addie Woicik, Linda~G Shapiro, Marian Blazes, Yue Wu, Cecilia~S Lee, Aaron~Y Lee, and Sheng Wang.
\newblock Octcube: a 3d foundation model for optical coherence tomography that improves cross-dataset, cross-disease, cross-device and cross-modality analysis.
\newblock {\em arXiv preprint arXiv:2408.11227}, 2024.

\bibitem{abbaspourazad2024wearable}
Salar Abbaspourazad, Anshuman Mishra, Joseph Futoma, Andrew~C Miller, and Ian Shapiro.
\newblock Wearable accelerometer foundation models for health via knowledge distillation.
\newblock {\em arXiv preprint arXiv:2412.11276}, 2024.

\bibitem{yang2023biot}
Chaoqi Yang, M.~Brandon Westover, and Jimeng Sun.
\newblock Biot: Biosignal transformer for cross-data learning in the wild.
\newblock In {\em NeurIPS 2023}, 2023.

\bibitem{oppenheim1999discrete}
Alan~V Oppenheim.
\newblock {\em Discrete-time signal processing}.
\newblock Pearson Education India, 1999.

\bibitem{daubechies1992ten}
Ingrid Daubechies.
\newblock {\em Ten lectures on wavelets}.
\newblock SIAM, 1992.

\bibitem{lee2025towards}
Simon~A Lee, Cyrus Tanade, Hao Zhou, Juhyeon Lee, Megha Thukral, Baiying Lu, and Sharanya~Arcot Desai.
\newblock Towards on-device foundation models for raw wearable signals.
\newblock In {\em NeurIPS 2025 Workshop on Learning from Time Series for Health}, 2025.

\bibitem{lee2025himae}
Simon~A Lee, Cyrus Tanade, Hao Zhou, Juhyeon Lee, Megha Thukral, Minji Han, Rachel Choi, Md~Sazzad~Hissain Khan, Baiying Lu, Migyeong Gwak, et~al.
\newblock Himae: Hierarchical masked autoencoders discover resolution-specific structure in wearable time series.
\newblock {\em arXiv preprint arXiv:2510.25785}, 2025.

\bibitem{mcdermott2025meds}
Matthew~BA McDermott, Justin Xu, Teya~S Bergamaschi, Hyewon Jeong, Simon~A Lee, Nassim Oufattole, Patrick Rockenschaub, Kamil{\.e} Stankevi{\v{c}}i{\=u}t{\.e}, Ethan Steinberg, Jimeng Sun, et~al.
\newblock Meds: Building models and tools in a reproducible health ai ecosystem.
\newblock In {\em Proceedings of the 31st ACM SIGKDD Conference on Knowledge Discovery and Data Mining V. 2}, pages 6243--6244, 2025.

\bibitem{kolo2024meds}
Aleksia Kolo, Chao Pang, Edward Choi, Ethan Steinberg, Hyewon Jeong, Jack Gallifant, Jason~A Fries, Jeffrey~N Chiang, Jungwoo Oh, Justin Xu, et~al.
\newblock Meds decentralized, extensible validation (meds-dev) benchmark: Establishing reproducibility and comparability in ml for health.
\newblock 2024.

\bibitem{wornow2024context}
Michael Wornow, Suhana Bedi, Miguel Angel~Fuentes Hernandez, Ethan Steinberg, Jason~Alan Fries, Christopher R{\'e}, Sanmi Koyejo, and Nigam~H Shah.
\newblock Context clues: Evaluating long context models for clinical prediction tasks on ehrs.
\newblock {\em arXiv preprint arXiv:2412.16178}, 2024.

\bibitem{odgaard2024core}
Mikkel Odgaard, Kiril~Vadimovic Klein, Sanne~M{\o}ller Thysen, Espen Jimenez-Solem, Martin Sillesen, and Mads Nielsen.
\newblock Core-behrt: A carefully optimized and rigorously evaluated behrt.
\newblock {\em arXiv preprint arXiv:2404.15201}, 2024.

\bibitem{shmatko2025learning}
Artem Shmatko, Alexander~Wolfgang Jung, Kumar Gaurav, S{\o}ren Brunak, Laust~Hvas Mortensen, Ewan Birney, Tom Fitzgerald, and Moritz Gerstung.
\newblock Learning the natural history of human disease with generative transformers.
\newblock {\em Nature}, 647(8088):248--256, 2025.

\bibitem{zhang2022tfc}
Xiang Zhang, Ziyuan Zhao, Theodoros Tsiligkaridis, and Marinka Zitnik.
\newblock Self-supervised contrastive pre-training for time series via time-frequency consistency.
\newblock In {\em NeurIPS}, 2022.

\bibitem{hu2024comprehensive}
Haigen Hu, Xiaoyuan Wang, Yan Zhang, Qi~Chen, and Qiu Guan.
\newblock A comprehensive survey on contrastive learning.
\newblock {\em Neurocomputing}, 610:128645, 2024.

\bibitem{jaiswal2020survey}
Ashish Jaiswal, Ashwin~Ramesh Babu, Mohammad~Zaki Zadeh, Debapriya Banerjee, and Fillia Makedon.
\newblock A survey on contrastive self-supervised learning.
\newblock {\em Technologies}, 9(1):2, 2020.

\bibitem{nguyen2022transformer}
Tung Nguyen and Aditya Grover.
\newblock Transformer neural processes: Uncertainty-aware meta learning via sequence modeling.
\newblock {\em arXiv preprint arXiv:2207.04179}, 2022.

\bibitem{ye2024exchangeable}
Naimeng Ye, Hanming Yang, Andrew Siah, and Hongseok Namkoong.
\newblock Exchangeable sequence models can naturally quantify uncertainty over latent concepts.
\newblock {\em arXiv preprint arXiv:2408.03307}, 2024.

\bibitem{ojha2025navigating}
Jaya Ojha, Oriana Presacan, Pedro G.~Lind, Eric Monteiro, and Anis Yazidi.
\newblock Navigating uncertainty: A user-perspective survey of trustworthiness of ai in healthcare.
\newblock {\em ACM Transactions on Computing for Healthcare}, 6(3):1--32, 2025.

\bibitem{hullermeier2021aleatoric}
Eyke H{\"u}llermeier and Willem Waegeman.
\newblock Aleatoric and epistemic uncertainty in machine learning: An introduction to concepts and methods.
\newblock {\em Machine learning}, 110(3):457--506, 2021.

\bibitem{lindenmeyer2025towards}
Adrian Lindenmeyer, Malte Blattmann, Stefan Franke, Thomas Neumuth, and Daniel Schneider.
\newblock Towards trustworthy ai in healthcare: Epistemic uncertainty estimation for clinical decision support.
\newblock {\em Journal of Personalized Medicine}, 15(2):58, 2025.

\bibitem{johnson2016mimic}
Alistair~EW Johnson, Tom~J Pollard, Lu~Shen, Li-wei~H Lehman, Mengling Feng, Mohammad Ghassemi, Benjamin Moody, Peter Szolovits, Leo Anthony~Celi, and Roger~G Mark.
\newblock Mimic-iii, a freely accessible critical care database.
\newblock {\em Scientific data}, 3(1):1--9, 2016.

\bibitem{johnson2023mimic}
Alistair~EW Johnson, Lucas Bulgarelli, Lu~Shen, Alvin Gayles, Ayad Shammout, Steven Horng, Tom~J Pollard, Sicheng Hao, Benjamin Moody, Brian Gow, et~al.
\newblock Mimic-iv, a freely accessible electronic health record dataset.
\newblock {\em Scientific data}, 10(1):1, 2023.

\end{thebibliography}

\end{document}